\documentclass[preprint,12pt]{elsarticle}

\usepackage{xurl}
\usepackage{hyperref}
\usepackage{amssymb}
\usepackage{amsmath}
\usepackage{xcolor}

\usepackage{booktabs}

\newcommand{\rot}[1]{\rotatebox{90}{#1}}

\journal{Knowledge-Based Systems}

\begin{document}

\begin{frontmatter}


\title{PREF-XAI: Preference-Based Personalized Rule Explanations of Black-Box Machine Learning Models}



\author{Salvatore Greco\fnref{unict}}
\ead{salgreco@unict.it}

\author{Jacek Karolczak\corref{cor1}\fnref{put}}
\ead{jacek.karolczak@cs.put.poznan.pl}

\author{Roman S{\l}owi{\'n}ski\fnref{put,IBS}}
\ead{roman.slowinski@cs.put.poznan.pl}

\author{Jerzy Stefanowski\fnref{put}}
\ead{jerzy.stefanowski@cs.put.poznan.pl}

\cortext[cor1]{Corresponding author}
\affiliation[unict]{organization={University of Catania, Department of Economics and Business}, 
            addressline={\\Corso Italia 55}, 
            city={Catania},
            postcode={95129}, 
            country={Italy}}
\affiliation[put]{organization={Poznan University of Technology, Institute of Computing Science}, 
            addressline={\\ Piotrowo 2}, 
            city={Pozna\'n},
            postcode={60-965}, 
            country={Poland}}

\affiliation[IBS]{organization={Systems Research Institute, Polish Academy of Sciences}, 
            addressline={\\ Newelska 6}, 
            city={Warsaw},
            postcode={01-447}, 
            country={Poland}}

\begin{abstract}
Explainable artificial intelligence (XAI) has predominantly focused on generating model-centric explanations that approximate the behavior of black-box models. However, such explanations often overlook a fundamental aspect of interpretability: different users require different explanations depending on their goals, preferences, and cognitive constraints. Although recent work has explored user-centric and personalized explanations, most existing approaches rely on heuristic adaptations or implicit user modeling, lacking a principled framework for representing and learning individual preferences. In this paper, we consider \emph{Preference-Based Explainable Artificial Intelligence (PREF-XAI)}, a novel perspective that reframes explanation as a preference-driven decision problem. Within PREF-XAI, explanations are not treated as fixed outputs, but as alternatives to be evaluated and selected according to user-specific criteria. In the PREF-XAI perspective, here we propose a methodology that combines rule-based explanations with formal preference learning. User preferences are elicited through a ranking of a small set of candidate explanations and modeled via an additive utility function inferred using robust ordinal regression. Experimental results on real-world datasets show that PREF-XAI can accurately reconstruct user preferences from limited feedback, identify highly relevant explanations, and discover novel explanatory rules not initially considered by the user. Beyond the proposed methodology, this work establishes a connection between XAI and preference learning, opening new directions for interactive and adaptive explanation systems.
\end{abstract}

\end{frontmatter}



\section{Introduction}
\label{sec:introduction}

Although modern ML models, such as neural networks or ensembles, achieve high predictive abilities in many tasks, they can be seen as ``black boxes'' because their complex internal logic and the reasons for individual decisions are incomprehensible to users. This has motivated the development of explainable AI (XAI), and many explanation methods based on different principles have been introduced \cite{guidotti2018xaisurvey,saeed2023survey}. However, they may provide explanations that are too complex or too generic, which do not adequately address the needs and preferences of individual users~\cite{byrne2023goodexplanations}.  

Building upon this motivation, in this paper, we introduce a \emph{Preference-Based Explainable Artificial Intelligence (PREF-XAI)} perspective, according to which explanations are not treated as mere outputs generated by an algorithm to approximate the behavior of a predictive model. Instead, they are viewed as alternative hypotheses that should be evaluated and selected according to user-specific priorities and expectations. 
Within this perspective, we propose an original methodology for personalized explanations of the predictions made by “black-box” machine learning (ML) classifiers.

Without compromising the high predictive accuracy of ``black-box'' ML models, we propose a personalized explainer that, on the one hand, provides explanatory information in an easily interpretable form and, on the other hand, tailors this information to user preferences expressed in a simple and intuitive way. 

An easily interpretable form of classification patterns is a set of ``if-then'' decision rules that relate a conjunction of elementary conditions on selected features to a suggested decision, e.g., ``if $at_i(\mathbf{x})\leqslant r_i,$ and $at_j(\mathbf{x})\geqslant r_j,\dots,$ and $at_k(\mathbf{x})=r_k,$ then $x \rightarrow$ class $c$'', where $i,j,\dots,k$ are indices of the selected features, $r_i,r_j,\dots,r_k$ are their threshold values discovered by induction from classification instances, and $\mathbf{x}$ is a classified instance. For a comprehensive review of the rules see, e.g.,~\cite{FGL12}. 

The strength of decision rules lies in modularity---each rule acts as a self-contained unit of knowledge---and the ability to help humans explicitly trace the relations between instance feature values and their class assignment. The symbolic representation of rules naturally corresponds to a high level of human comprehensibility \cite{MekonnenLuca2024,letham2015interpretable}.

Rules can serve as interpretable proxies for black-box models in two ways: as \textit{global explanations}, a set of rules approximates the overall behavior of a black-box model, providing a surrogate representation of its decision logic; or as \textit{local explanations}, individual rules explain why a specific prediction was made for a particular instance \cite{bodria2023benchmarking}. In this study, we focus specifically on the latter approach.

Despite their high explanatory value, rule-based XAI methods focus primarily on model-centric properties. For instance, methods often extract patterns based on feature relevance~\citep{macha2022rulexai} or fuse rule induction with XAI-based feature importance to strictly mimic the black box~\citep{kozielski2025fusion}. While these approaches improve the mathematical consistency of the explanation, they operate independently of the user's needs. To address this, some methods attempt a more user-centric design~\citep{kaplan2024usercentric}, such as the framework proposed in~\citep{nimmy2023brb} which aims to enhance explanation trustworthiness.

However, these approaches still fail to account for subjective user preferences. Rather than actively tailoring the output to the individual, they typically generate multi-view explanations. This merely shifts the cognitive burden onto the user, requiring them to manually evaluate and select their preferred option.

Conversely, existing approaches to personalized explanations predominantly focus on Large Language Models (LLMs) and conversational agents ~\citep{lubos2024llmpersonal,song2025llmpersonal}. However, their practical usefulness and the presentation of answers that are well-grounded in facts remain unsatisfactory due to low fidelity and faithfulness~\citep{mayne2025llmboundries,turpin2023llmunfaithful}.

We now explain the general methodology presented in this paper. Let's assume that the ``black-box'' ML classifier is a Multilayer Perceptron (MLP). The MLP is first trained on a training set of labeled instances. This follows the logic of methods like LORE~\cite{guidotti2018lore}, where the original labels of training instances are replaced with those produced by the trained MLP to ensure that the induced rules accurately reflect the model's decision logic. 

LORE~\cite{guidotti2018lore} and Anchor~\cite{ribeiro2018anchor} are designed to identify a single post hoc rule for a given classified instance. That rule is derived from artificially generated examples in the instance’s neighborhood, and its conditions are selected through the automatic optimization of rule-evaluation metrics, rather than in a way that adequately reflects the user’s preferences. Moreover, because this rule usually involves only a limited number of conditions, it provides only partial insight into the circumstances underlying the black box’s decision-making.

Our methodology first constructs a comprehensive surrogate model by inducing a global set $\mathcal{R}$ of ``if-then'' decision rules from the modified training set.

For an unseen instance $\mathbf{x}$, the MLP is first used to produce a classification. The rules from the subset $\mathcal{R}_{\mathbf{x}} \subset \mathcal{R}$ that match $\mathbf{x}$ and its MLP-generated label are then used to explain the MLP’s decision. 

Remark that the set $\mathcal{R}_\mathbf{x}$ may, in general, be large, making the explanation too vague. For this reason, it is reasonable to extract from this set a smaller subset of rules that comply with the user’s preferences. User preferences often center on specific features that they consider to be most relevant for understanding model behavior. User preferences may also extend to the attractiveness measures evaluating rules in relation to the training data, such as rule support, confidence, or their Bayesian confirmation \cite{greco2016measures}, as well as the complexity of the rule (the number of elementary conditions). When the set of rules $\mathcal{R}_\mathbf{x}$ is ranked according to the user’s preferences, the resulting explanation becomes personalized, which is precisely the goal of our methodology.

Two questions naturally arise: (i) which preference model is appropriate for ranking the rules in $\mathcal{R}_\mathbf{x}$, and (ii) what preference information can reasonably be obtained from the user to construct such a model? The choice of preference models used in constructive multiple criteria decision aiding (MCDA) and machine preference learning (PL) has recently been characterized in \citep{HullermeierSlowinski2024}. In particular, a popular model is a utility function that aggregates multiple criteria evaluations of an instance into a real-valued score. In our approach, to address (i), we adopt an additive utility function model expressed as a weighted sum of the criteria used to evaluate rules. The evaluation criteria include the presence/absence of each feature in a rule (coded as 1/0), the rule support, the rule confirmation, and the rule complexity. With respect to (ii), we follow the prevailing trend toward holistic preference information elicitation instead of the direct elicitation of utility function weights. The holistic preference information is provided by the user as a ranking over a limited subset of selected rules, denoted by $\mathcal{R}_\mathbf{x}^u \subset \mathcal{R}_\mathbf{x}$. As argued in \citep{greco2024fifty}, contemporary MCDA approaches increasingly rely on holistic elicitation, as directly specifying preference model parameters places an excessive cognitive burden on users. 

The construction of the utility preference model from the preference ranking of a limited number of rules selected from $\mathcal{R}_\mathbf{x}$ is performed using the robust ordinal regression (ROR) method described in \citep{CorrenteEtAl2013,GrecoMousseauSlowinski2008}. ROR identifies the whole set of instances of the assumed utility function that reproduces the ranking of the selected rules provided by the user. By the whole set of instances, we mean all vectors of weights of the utility function that are compatible with the user’s preference ranking. These vectors form a convex polyhedron (simplex). 

Any utility function constructed from a weight vector belonging to this polyhedron induces the same ranking of the rules from $\mathcal{R}_\mathbf{x}^u$ as specified by the user, while the order relations among the remaining rules in $\mathcal{R}_\mathbf{x}$ may vary. To obtain a single representative ranking of the rules in $\mathcal{R}_\mathbf{x}$, we consider two approaches. The first consists of identifying a compatible weight vector that maximizes the minimum utility difference between consecutive rules $r_i,r_j \in \mathcal{R}_\mathbf{x}^u$, and then ranking all rules in $\mathcal{R}_\mathbf{x}$ according to the corresponding utility function. The second strategy explores the entire volume of the compatible weight space by performing a well-distributed random sampling of a large number of compatible weight vectors (e.g., 10,000). Assuming a uniform probability distribution, this sampling is executed using the ``hit-and-run'' algorithm \citep{HaR}. From this extensive sample, we extract the final preference representations in two distinct ways: by averaging the sampled vectors to establish a single, robust geometric centroid (H\&R$^C$) used to rank all rules; or by applying the First Rules approach (H\&$^{FR}$), which evaluates rule rankings across all sampled utility functions to yield an unordered subset of rules that occupy the top position at least once. The final outcomes resulting from these approaches are presented to the user, representing the most preferred personalized explanations of the decision concerning instance $\mathbf{x}$.

To the best of our knowledge, personalizing rule-based explanations of black-box ML models  via formal preference modeling has not been previously investigated in the XAI literature. It employs a set of human-interpretable ``if-then'' decision rules to explain an MLP classification decision concerning an instance $\mathbf{x}$, with the rules ranked according to the user’s preferences expressed via a preference ranking of a small number of exemplary rules matching $\mathbf{x}$. If the obtained ranking is not satisfactory, the procedure can be iterated by acquiring additional preference information from the user.

The remainder of the paper is organized as follows. Section 2 details the proposed methodology, starting from the general workflow and covering the MLP learning phase, rule induction, robust ordinal regression applied to the user’s preference information, and stochastic multicriteria decision analysis. Section 3 provides an illustrative example of the methodology and presents computational experiments conducted on three real-world tabular datasets. After describing the objectives of the experiments, we report the experimental setup, followed by the results and their discussion. Section 4 concludes the paper and outlines directions for future research. 

\section{Methodology}
\label{sec:methodology}

\subsection{Workflow}
The proposed methodology relies on two primary prerequisites. First, the availability of a black-box ML classifier, previously trained on a dataset, is assumed. Second, a comprehensive set of ``if-then'' decision rules capable of explaining the model's decisions is required.

To construct a personalized explanatory model for a specific instance, the workflow proceeds through a sequence of interactive steps. Initially, the subset of rules that cover the given instance and match the prediction of the black-box model is identified. Because this covering set is often too large for direct human evaluation, the user is asked to select and rank a small, manageable subset of these rules.

Using the provided reference ranking, an ordinal regression problem is formulated and solved. This step infers the weights of a user-specific preference model, which takes the form of a weighted value function called the Preference-based Rule Utility Score (PRUS). Once established, the PRUS is applied to evaluate and rank the entire set of covering rules. The final comprehensive ranking is then presented to the user, with the most highly evaluated rules serving as the personalized explanations. If the resulting explanations are not satisfactory, the procedure can be repeated, allowing the preference model to be iteratively refined through a new reference ranking.

\subsection{Mathematical Formulation}
\label{sec:mathematical-formulation}

\subsubsection{Notation}
A \emph{dataset} $\mathcal{S}$ is a collection of $n$ instances, $\mathcal{S} = \{(\mathbf{x}_i, y_i)\}_{i=1}^{n}$, where each $\mathbf{x}_i \in \mathbb{X}^d$ is a $d$-dimensional feature vector and $y_i \in \mathcal{Y}$ is the ground-truth label. A machine learning model is a function $h: \mathbb{X}^d \rightarrow \mathcal{Y}$ that maps an input instance to a predicted class $\hat{y} = h(\mathbf{x})$. While such models are often black boxes, their local decision logic can be approximated using symbolic rules.

We use \emph{rules} as a post-hoc explanation technique. A rule is formally defined as a tuple $r = (\mathcal{C}, c)$, where $\mathcal{C}$ is a conjunction of elementary conditions (the antecedent) and $c$ is the model's predicted decision (the consequent). A single elementary condition is expressed as $(x^{(j)} \in v)$, where $x^{(j)}$ is the $j$-th feature of $\mathbf{x}$ and $v$ is a discrete category or interval. An instance $\mathbf{x}$ is \emph{covered} by a rule $r$ if it satisfies all conditions in $\mathcal{C}$. We assume a universal set of rules $\mathcal{R}$ such that for any instance $\mathbf{x}$, there exists a subset of valid rules $\mathcal{R}_{\mathbf{x}} \subseteq \mathcal{R}$ matching $\mathbf{x}$. For every valid rule $r = (\mathcal{C}, c) \in \mathcal{R}_{\mathbf{x}}$, two conditions hold: $\mathcal{C}$ covers $\mathbf{x}$, and the consequent matches the model prediction ($c = h(\mathbf{x})$).

The number of valid rules, $|\mathcal{R}_{\mathbf{x}}|$, varies across domains and often grows exponentially. This combinatorial explosion frequently occurs in datasets with many features or strong feature correlations, where multiple distinct antecedents describe the same local region, necessitating a dense set of rules for an accurate local approximation.

\subsubsection{Rule Induction}

To induce the candidate rule set $\mathcal{R}$, we adapt Classification Based on Association (CBA)---an association rule mining technique to generate informative and descriptive classification rules \cite{FGL12,Liu1998}. To formulate the elementary conditions defined above, continuous features are first discretized into intervals using quantile-based binning. Next, all categorical and discretized features are one-hot encoded. We then apply the Apriori algorithm \citep{agrawal1994apriori} to identify frequent itemsets that meet a minimum support threshold. From these itemsets, we induce association rules, filtering the output to retain only predictive rules corresponding to target classes.

To quantitatively evaluate the induced rules, we use three primary criteria for each rule $r = (\mathcal{C}, c)$:
\begin{itemize}
    \item \textbf{Support:} The fraction of instances in the dataset that satisfy both the antecedent $\mathcal{C}$ and the consequent $c$. This quantity estimates the probability $P(\mathcal{C} \land c)$ that an instance from the dataset is matched by the rule. Let $N_{total}$ be the total number of instances in the dataset, and $n(\mathcal{C} \land c)$ be the number of instances matching the full rule. Support is formulated as:
    \begin{equation}
        \text{Support}(r) = P(\mathcal{C} \land c) = \frac{n(\mathcal{C} \land c)}{N_{total}}\,.
    \end{equation}
    
    \item \textbf{Confirmation:} A Bayesian confirmation measure quantifying the degree of evidential support that the antecedent $\mathcal{C}$ provides for the consequent $c$. Specifically, we utilize Nozick's $N$ measure, which evaluates the difference between the probability of the antecedent given the presence versus the probability given the absence of the consequent \citep{greco2016measures}:
    \begin{equation}
        \text{Confirmation}(r) = N(c, \mathcal{C}) = P(\mathcal{C}|c) - P(\mathcal{C}|\neg c)\,.
    \end{equation}
    
    \item \textbf{Complexity:} The structural length of the rule, defined as the number of elementary conditions contained in the rule antecedent:
    \begin{equation}
        \text{Complexity}(r) = |\mathcal{C}|\,.
    \end{equation}
    Because a rule can contain at most one condition per feature, the maximum possible complexity is bounded by the total number of features $d$.
\end{itemize}

\subsubsection{Ordinal Regression}

To recap, from the set of rules covering the instance currently being classified, the user selects a reference set and ranks it according to their preferences.

We define a value function to represent the user’s preferences, termed the Preference-based Rule Utility Score (PRUS). An additive value model is adopted, assuming mutual preferential independence among the marginal utilities. In addition to Support, Confirmation, and Complexity, the marginal utilities also account for the presence of specific elementary conditions (Cond($r$)) in the evaluated rule. The overall utility of a rule $r$ is expressed as the following weighted sum:
\begin{equation}
    \label{eq:prus}
    \begin{aligned}
        PRUS(r) =\;& \sum_{j \in Cond(r)} w_j + w_{supp} \times \text{Support}(r) \\ 
        &+ w_{conf} \times \text{Confirmation}(r) \\
        &- w_{comp} \times \frac{|\text{Complexity}(r)|}{\text{max possible } |\text{Complexity}(r)|}\,,
    \end{aligned}
\end{equation}
with nonnegative weights summing to one.
Notice that rule complexity is penalized in the formulation, reflecting a general user preference for shorter, more interpretable explanations.

To determine the weights of PRUS, we formulate an ordinal regression problem below. The objective is to reproduce the reference ranking of selected rules provided by the user using PRUS. The user ranks a set of rules from most interesting ($r_1$) to least interesting ($r_h$). We enforce linear constraints to maintain this exact order. The scalar $\epsilon$ acts as a strictly positive margin, ensuring that a strictly preferred rule has a tangibly higher utility score than the subsequent one. The sum of all weights is normalized to 1, and all weights must be non-negative.

\begin{equation}
\label{eq:OR}
    \begin{aligned}
        & \text{Maximize} & & \epsilon \\
        & \text{Subject to} & & PRUS(r_k) \ge PRUS(r_{k+1}) + \epsilon, \quad \forall k \in \{1, \dots, h-1\} \\
        & & & \sum_{j} w_j + w_{supp} + w_{conf} + w_{comp} = 1 \\
        & & & w_j, w_{supp}, w_{conf}, w_{comp} \ge 0\,.
    \end{aligned}
\end{equation}

The linear constraints of problem (\ref{eq:OR}) define a feasible convex polyhedron in the multidimensional weight space. The PRUS function with any weight vector within this polyhedron perfectly reconstructs the user's reference ranking. We refer to these weight vectors as compatible.

\subsubsection{Robust Ordinal Regression with Hit-and-Run sampling}
Because, in general, many compatible weight vectors satisfy the ordinal regression constraints of problem (\ref{eq:OR}), selecting a single arbitrary vector may not reliably represent the user's true preference system. To address the parameter uncertainty of PRUS, we transition to Robust Ordinal Regression (ROR) \citep{GrecoMousseauSlowinski2008}. We will apply two different strategies to resolve the ambiguity of the PRUS weight vectors compatible with the reference ranking.

\paragraph{Compatible weight vector maximizing discrimination (Max $\epsilon$)}

This strategy consists of solving the linear programming problem (\ref{eq:OR}). It identifies a compatible weight vector that maximizes the minimal utility difference ($\epsilon$) between consecutive rules in the user’s reference ranking. Geometrically speaking, it finds a solution deep within the interior of the polyhedron, representing a ``max–min margin'' preference model that is highly robust to minor perturbations~\cite{CorrenteEtAl2013}.

\paragraph{Hit-and-run sampling (H\&R$^C$ and H\&R$^{FR}$)}

Instead of relying on a single max–min margin weight vector, this approach explores the entire volume of the compatible weight space. According to this strategy, a modified linear program is first solved to obtain an initial feasible solution. In this formulation, all weights are required to be strictly positive by imposing a lower bound of $\epsilon$. The linear program takes the following form:

\begin{equation}
\label{eq:OR2}
    \begin{aligned}
        & \text{Maximize} & & \epsilon \\
        & \text{Subject to} & & PRUS(r_k) \ge PRUS(r_{k+1}), \quad \forall k \in \{1, \dots, h-1\} \\
        & & & \sum_{j} w_j + w_{supp} + w_{conf} + w_{comp} = 1 \\
        & & & w_j, w_{supp}, w_{conf}, w_{comp} \ge \epsilon\,.
    \end{aligned}
\end{equation}

Starting from this initial interior point, we apply the ``hit-and-run'' algorithm described in \citep{HaR}. This is a Markov Chain Monte Carlo (MCMC) method that uniformly samples the multidimensional convex polyhedron of compatible weight vectors. These samples are very numerous and, in our case, contain 10,000 compatible weight vectors. From these samples, we derive two distinct solutions:
\begin{itemize}
    \item \textbf{Geometric centroid (H\&R$^C$):} Obtained by averaging the sampled vectors to produce a single, robust weight vector representing a ``central'' preference model.
    \item \textbf{First Rules approach (H\&R$^{FR}$):} This approach does not aggregate the weights; instead, it evaluates rule rankings across all sampled utility functions and selects the rules that occupy the top position at least once, thereby yielding an unordered subset of rules.
\end{itemize}

\section{Computational Experiments}
This section evaluates the proposed methodology through computational experiments. We begin by defining the research questions and the experimental setup, followed by the evaluation metrics. We then provide an illustrative example to build intuition. Finally, we conduct experiments on three datasets to address the research questions.

\subsection{Aims}

The computational experiments evaluate the effectiveness and reliability of the proposed personalized explanation methodology. Specifically, the experiments are designed to answer the following research questions:
\begin{itemize}
    \item[\textbf{RQ1:}] To what extent do the top-ranked rules identified by the algorithm align with the true most preferred rules of the user?
    \item[\textbf{RQ2:}] How accurately do robust ordinal regression models reproduce the underlying Preference-based Rule Utility Score (PRUS) of the user from a limited reference ranking?
    \item[\textbf{RQ3:}] What is the capacity of the proposed algorithm to discover novel, highly relevant rules that are absent from the initial reference set?
    \item[\textbf{RQ4:}] Which optimization strategy for deriving a compatible weight vector for PRUS yields better performance across the evaluation metrics: the sample of compatible weight vectors computed via hit-and-run (H\&R$^C$), or the compatible weight vector that maximizes discrimination (Max~$\epsilon$)?
\end{itemize}

\subsection{Experimental Setup}

\subsubsection{Data}
\label{sec:data}

The proposed method is evaluated on three real-world tabular datasets: Churn (Banking)\footnote{Sourced from: \href{https://www.kaggle.com/datasets/shrutimechlearn/churn-modelling}{https://www.kaggle.com/datasets/shrutimechlearn/churn-modelling}}, Churn (Telecom)\footnote{Sourced from: \href{https://www.kaggle.com/datasets/becksddf/churn-in-telecoms-dataset}{https://www.kaggle.com/datasets/becksddf/churn-in-telecoms-dataset}}, and HELOC\footnote{Sourced from: \href{https://www.kaggle.com/datasets/averkiyoliabev/home-equity-line-of-creditheloc}{https://www.kaggle.com/datasets/averkiyoliabev/home-equity-line-of-creditheloc}}. These datasets were selected because they represent high-stakes decision-making environments, customer retention, and credit risk assessment, where the alignment between model logic and human preference is critical for trust and regulatory compliance. Furthermore, they provide structural diversity across sample size, dimensionality, and class distribution. Prior to training, highly correlated and non-discriminative features (e.g., customer IDs, phone numbers) and private features (e.g., surnames) were removed to ensure that the model $h$ relies solely on generalizable behavioral and financial patterns.

\subsubsection{Training ML Model}

To create a black-box classifier $h$, we trained the Multilayer Perceptron (MLP) model\footnote{All trainings were conducted using the PyTorch framework: \href{https://pytorch.org}{https://pytorch.org}}. The network architecture of $h$ comprises two hidden layers, each containing 32 neurons. This configuration was chosen because empirical tests showed that larger models did not yield further performance gains, and the current setup successfully matches the predictive performance reported in the literature. To effectively handle categorical features, we utilize learnable embeddings of size 2 for each categorical input. The Rectified Linear Unit (ReLU) activation function \citep{goodfellow2016deeplearningbook} is applied after each hidden layer. The model parameters are optimized using AdamW over the dataset $\mathcal{S}$ to minimize the Binary Cross-Entropy loss ($\mathcal{L}_{BCE}$)~\cite{goodfellow2016deeplearningbook}. To mitigate the impact of class imbalance within the dataset $\mathcal{S}$, we apply specific class weights within the loss function and refine the decision threshold used by $h$ to assign positive labels. For regularization, a weight decay of $1 \times 10^{-5}$ is applied during optimization.

We perform hyperparameter tuning to ensure that the predictive performance of $h$ is comparable to other models reported in the literature for these tasks. The resulting predictive performance, measured in terms of accuracy and balanced accuracy (considered for class imbalances) via the 10-fold cross-validation, is reported in Tab.~\ref{tab:accuracy}. All specific hyperparameters determined through this process, including learning rates, batch sizes, class weights, and decision thresholds for each dataset, are detailed in Tab.~\ref{tab:hparams}.

\begin{table}[tbp]
    \centering
    \begin{tabular}{l c r}
        \toprule
        Dataset & Accuracy & Balanced Accuracy \\ 
        \midrule
        Churn (Banking) & $0.864\,(\pm 0.009)$ & $0.738\,(\pm 0.013)$ \\
        Churn (Telecom)  & $0.958\,(\pm 0.010)$ & $0.899\,(\pm 0.034)$ \\
        HELOC           & $0.767\,(\pm 0.013)$ & $0.766\,(\pm 0.013)$ \\
        \bottomrule
    \end{tabular}

    \caption{\textbf{Predictive Performance of the MLP Classifier:} Mean accuracy and balanced accuracy scores obtained via 10-fold cross-validation across the datasets. Standard deviations are reported in parentheses to indicate performance stability over the folds.}
    \label{tab:accuracy}
\end{table}

\begin{table}[tbp]
    \centering
    \begin{tabular}{l c c c}
        \toprule
        Hyperparameter & Churn (Banking) &  Churn (Telecom) & HELOC \\ 
        \midrule
        Learning Rate & $1 \times 10^{-3}$ & $3 \times 10^{-3}$ & $1 \times 10^{-3}$ \\
        Batch Size  & $1024$ & $512$ & $512$\\
        $\mathcal{L}_{BCE}$ Class Weights & $[1.0, 3.5]$ & $[1.0, 9.0]$ & $[1.0, 1.0]$ \\
        Threshold & $0.9$ & $0.5$ & $0.5$ \\
        \bottomrule
    \end{tabular}
    \caption{\textbf{Model Hyperparameters:} Specific training configuration values used for MLP across the different datasets. These parameters were selected to optimize predictive performance and handle class imbalance.}   
    \label{tab:hparams}
\end{table}

\subsubsection{Rule Generation Setup}

\begin{figure}[tbp]
    \centering
    \includegraphics[width=0.5\linewidth]{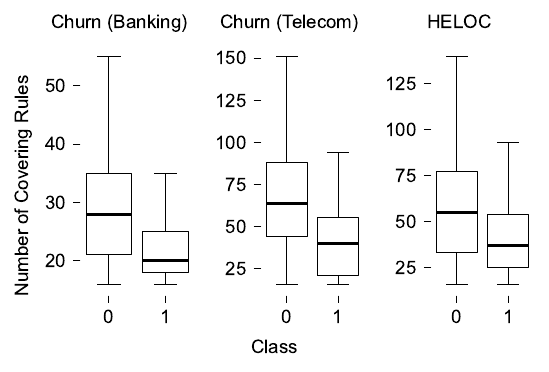}
    \caption{\textbf{Number of Covering Rules:} The number of decision rules covering an individual instance, stratified by the predicted class for each dataset. This illustrates the variable size of the initial rule set $\mathcal{R}_{\mathbf{x}}$, highlighting the need to generate personalized explanations.}
    \label{fig:number-of-covering-rules}
\end{figure}

Table~\ref{tab:apriori-parameters} details the number of bins (discretization intervals), minimum support, confirmation, and confidence thresholds used for each dataset. Crucially, we enforce a confidence threshold of $1.0$. This parameter choice is deliberate: it guarantees perfect fidelity, ensuring that the induced rules act as an exact logical representation of the black-box model's predictions rather than a lossy approximation. Specific values of the remaining criteria were chosen to ensure that the rule induction process remains computationally feasible on a standard laptop. Accordingly, all experiments reported in this study were executed on a machine equipped with an Apple M2 processor and 16 GB of RAM. Figure~\ref{fig:number-of-covering-rules} reports the resulting number of valid rules covering each instance across the datasets, stratified by target class.

\begin{table}[tbp]
    \centering
    \begin{tabular}{l c c c c}
        \toprule
        Dataset & Min. Supp. & Min. Confirm. & Min. Conf. & Num. Bins \\
        \midrule
        Churn (Banking) & 1 & 0.0 & 1.0 & 4 \\
        Churn (Telecom) & 4 & 0.0 & 1.0 & 3 \\
        HELOC & 3 & 0.0  & 1.0 & 3 \\
        \bottomrule
    \end{tabular}
    \caption{\textbf{Rule Induction Parameters:} Specific parameter values used in the Apriori algorithm to induce the candidate set of decision rules across the evaluated datasets. The parameters define the thresholds for minimum support, confirmation, and confidence, as well as the number of bins used for continuous feature discretization. Repository links for these datasets are provided in Section~\ref{sec:data}.}
    \label{tab:apriori-parameters}
\end{table} 

\subsubsection{Simulated User}
To conduct large-scale computational experiments, we employ an automated procedure to simulate the behavior of a human user. For a given instance $\mathbf{x}$, we identify the set of all covering rules $\mathcal{R}_{\mathbf{x}}$. The simulated user is defined by a randomly generated true utility function $U_{\text{true}}$, where each rule evaluation criterion, including rule conditions, support, confirmation, and complexity, is assigned a weight $w_i$ drawn from a uniform distribution $\mathcal{U}(0,1)$ and subsequently normalized such that $\sum w_i = 1$. By drawing weights from a uniform distribution, we simulate a broad spectrum of potential preference profiles. This approach ensures that the methodology is evaluated for an arbitrary user, avoiding a priori biases toward specific rule characteristics or evaluation criteria. The utility function $U_{\text{true}}(r)$ has the same mathematical form as PRUS (\ref{eq:prus}). Using this ground-truth utility function, we calculate a true utility score $U_{\text{true}}(r)$ for every rule $r \in \mathcal{R}_{\mathbf{x}}$. To emulate the preference elicitation process, we randomly sample a small subset of rules $\mathcal{R}_{\mathbf{x}}^u \subset \mathcal{R}_{\mathbf{x}}$ with a cardinality $k$. The simulated user ranks this subset in descending order of the scores $U_{\text{true}}(r)$, generating the reference ranking. Of course, the ordinal regression does not have access to the user’s utility function $U_{\text{true}}(r)$ and its weights.

\subsection{Evaluation Metrics}

To quantify the performance of the proposed ordinal regression strategies with respect to the simulated user's rankings, we employ the following measures. 

\paragraph{Ranking Fidelity}
To assess ranking fidelity, we calculate Kendall's rank correlation coefficient $\tau$ between the full ranking of all covering rules generated by the algorithm and the ground-truth ranking derived from $U_{\text{true}}$. For two rankings $u$ and $v$ of $n$ rules, Kendall's $\tau$ is defined as:
\begin{equation}
    \tau = \frac{2}{n(n-1)} \sum_{i<j} \text{sgn}(u_i - u_j)\text{sgn}(v_i - v_j)\,,
\end{equation}
where $u_i$ and $v_i$ represent the ranks assigned to rule $i$, with $i=1,\ldots,n$. This measure evaluates the degree of agreement between the rule orderings generated by PRUS($r$) and $U_{\text{true}}(r)$. When it equals 1, the two rankings are identical; when it equals $-1$, they are completely reversed; and when it equals 0, there is no correlation.

\paragraph{Parameter Recovery}
We measure parameter recovery by calculating Kendall's $\tau$ between the weight vector derived from the ordinal regression and the ground-truth weights of the simulated user. This evaluates how closely the learned parameters approximate the true underlying preferences in terms of utility function weights.

\paragraph{Top-$k$ Rule Set Similarity}
We evaluate top-$k$ rule set similarity using the Jaccard index to measure the overlap between the sets of top-ranked rules in both rankings. Let $A_k$ denote the set of the top $k$ rules selected by the algorithm and $B_k$ denote the true top $k$ rules preferred by the user. The Jaccard index is formulated as:
\begin{equation}
    J(A_k, B_k) = \frac{|A_k \cap B_k|}{|A_k \cup B_k|}\,.
\end{equation}
Note that the alternative First Rules strategy (H\&R$^{FR}$) is evaluated exclusively using this criterion. Because H\&R$^{FR}$ directly aggregates the most frequent top-ranked rules across sampled functions rather than deriving a single representative weight vector or full ranking, correlation-based metrics such as parameter recovery and ranking fidelity cannot be applied.

\paragraph{Rule Discovery}
We quantify rule discovery through two specific metrics. First, we calculate the number of newly discovered rules, defined as the total count of rules ranked algorithmically by PRUS above the highest-ranked rule from the initial reference set $\mathcal{R}_{\mathbf{x}}^u$. Second, we measure the number of new rules within the top-5 positions, which counts exactly how many rules occupying the top $5$ ranks of the final output were not elements of the reference set $\mathcal{R}_{\mathbf{x}}^u$. These metrics highlight the method's ability to recommend valuable unseen patterns that align with the learned preferences.

\paragraph{Statistical Significance Testing}
We apply the non-parametric Wilcoxon signed-rank test~\cite{rey2011wilcoxon} to evaluate the statistical significance of performance differences between the optimization methods. Results for the paired metrics, such as Kendall's $\tau$ and the Jaccard index, are reported using common $p$-value thresholds (i.e., $p < 0.05$, $p < 0.01$, and $p < 0.001$).

\subsection{Illustrative Example}

\begin{table}[tbp]
    \centering
    \begin{tabular}{l c}
        \toprule
        Feature & Value \\
        \midrule
        Clump Thickness               & 3 \\
        Uniformity of Cell Size       & 1 \\
        Uniformity of Cell Shape      & 1 \\
        Marginal Adhesion             & 1 \\
        Single Epithelial Cell Size   & 2 \\
        Bare Nuclei                   & 1 \\
        Bland Chromatin               & 1 \\
        Normal Nucleoli               & 1 \\
        Mitoses                       & 1 \\
        \midrule
        Class                         & benign (0) \\
        \bottomrule
    \end{tabular}
    \caption{Feature values for the illustrative instance $\mathbf{x}$ from the Wisconsin Original dataset.}
    \label{tab:toy-example-instance}
\end{table}

Before proceeding to the mass experiments, we provide a pedagogical example showing how the proposed method will be executed step by step, using a small example based on the Wisconsin Original dataset\footnote{Sourced from: \href{https://www.kaggle.com/datasets/mariolisboa/breast-cancer-wisconsin-original-data-set}{https://www.kaggle.com/datasets/mariolisboa/breast-cancer-wisconsin-original-data-set}}. This dataset has been used both to train an MLP and to generate ``if–then'' decision rules. A new instance $\mathbf{x}$, shown in Table~\ref{tab:toy-example-instance}, has been assigned by the trained MLP to class benign (0). Instance $\mathbf{x}$ was also covered by 16 decision rules indicating the same class. It was deliberately selected to keep the example concise and easily readable, as other instances often trigger much larger sets of matching rules. Table~\ref{tab:toy-example} presents these 16 rules, which explain the MLP’s classification through elementary conditions that match the instance's specific feature values.

To simulate the user's decision-making process, we define a ground-truth utility function $U_{\text{true}}$ as a weighted sum of the twelve elements (corresponding to features occurring in rule conditions and rule evaluation criteria). The vector of weights, randomly generated and normalized, is as follows:
\begin{equation}
    \mathbf{w}_{U_{\text{true}}} =
        \left(
        \begin{array}{rl}
            w_{\text{Thick}} =0.060\\
            w_{\text{Size}} =0.153\\
            w_{\text{Shape}} =0.118\\
            w_{\text{Adh}} =0.096
        \end{array}\, 
        \begin{array}{rl}
            w_{\text{Epith}} =0.025\\
            w_{\text{Bare}} =0.025\\
            w_{\text{Chrom}} =0.009\\
            w_{\text{Nuc}} =0.139 
        \end{array}\,
        \begin{array}{rl}
            w_{\text{Mit}} =0.097 \\
            w_{\text{supp}} =0.114 \\
            w_{\text{conf }} =0.003 \\
            w_{\text{comp}} =0.157
        \end{array} \right)\,
    \end{equation}

Guided by these underlying preferences (modeled using the above-mentioned utility function), the simulated user ranks a small subset of 10 rules, indicated in Table~\ref{tab:toy-example} by the $\star$ symbol, as follows:
\begin{equation}
    \label{eq:reference_ranking}
    r_3 \rightarrow r_6 \rightarrow r_9 \rightarrow r_7 \rightarrow r_{12} \rightarrow r_{10} \rightarrow r_8 \rightarrow r_{14} \rightarrow r_{0} \rightarrow r_{2}\,.
\end{equation}

The user is asked to provide a short ranking of rules. However, when the set of compatible rules is large---for example, comprising 50 rules---evaluating the entire set becomes cognitively too demanding. Consequently, the initial reference ranking presented above encompasses only a manageable subset of all rules covering the explained instance. 

Using the reference ranking of 10 rules, the robust ordinal regression algorithm computes a compatible weight vector that maximizes discrimination among the ranked rules (Max~$\epsilon$). The compatible PRUS weights have the following values:

\begin{equation}
    \label{eq:toy_prus_weights}
    \mathbf{w}_{\text{PRUS}} =
    \left(
    \begin{array}{rl}
        w_{\text{Thick}} =0.038\\
        w_{\text{Size}} =0.227\\
        w_{\text{Shape}} =0.143\\
        w_{\text{Adh}} =0.124
    \end{array}\, 
    \begin{array}{rl}
        w_{\text{Epith}} =0.000\\
        w_{\text{Bare}} =0.000\\
        w_{\text{Chrom}} =0.000\\
        w_{\text{Nuc}} =0.192 
    \end{array}\,
    \begin{array}{rl}
        w_{\text{Mit}} =0.095 \\
        w_{\text{supp}} =0.000 \\
        w_{\text{conf }} =0.180 \\
        w_{\text{comp}} =0.000
    \end{array} \right)\,
\end{equation}

The inferred PRUS weights provide insight into the specific priorities extracted from the user's reference ranking. Regarding the rule evaluation criteria, PRUS identifies a strong preference for the confirmation measure ($w_{\text{conf}} = 0.180$), while assigning zero weight to rule support and complexity. Although the ground-truth utility function $U_{\text{true}}$ indicates that the simulated user actually prioritized support ($0.114$) over confirmation ($0.003$), this discrepancy is not problematic; support and confirmation frequently exhibit a strong empirical correlation within induced rule sets, allowing the ordinal regression to effectively approximate the user's overall rule ranking by leveraging a correlated substitute criterion.

Concerning the elementary conditions, the PRUS weights reveal a strong preference for explanations that include Uniformity of Cell Size ($0.227$), Normal Nucleoli ($0.192$), and Uniformity of Cell Shape ($0.143$). This is closely consistent with the underlying ground-truth preferences in $U_{\text{true}}$, where the values for these specific feature weights are $0.153$, $0.192$, and $0.143$, respectively.

This substantial alignment between the inferred and actual weight vectors is reflected by the parameter recovery, quantitatively expressed by Kendall's $\tau$ correlation of 0.2961  between the recovered PRUS weight vector and the ground-truth $U_{\text{true}}$ weight vector.

Then, PRUS function (\ref{eq:prus}) with the derived weight vector (see Equation~\ref{eq:toy_prus_weights}) is used to score and rank all 16 decision rules. The right column of Table~\ref{tab:toy-example} shows the scores of all 16 rules provided by PRUS. Based on these scores, the rules are ranked as follows:
\begin{equation}
    \label{eq:PRUS_ranking}
    \begin{gathered}
        r_5 \rightarrow r_{3} \rightarrow r_{11} \rightarrow r_6 \rightarrow r_9 \rightarrow r_7 \rightarrow r_{12} \rightarrow r_{4} \\
        \rightarrow r_{10} \rightarrow r_{8} \rightarrow r_{13} \rightarrow r_{14} \rightarrow r_{0} \rightarrow r_{1} \rightarrow r_{2}\,.
    \end{gathered}
\end{equation}

In this illustrative example, the ground-truth ranking of all 16 rules provided by the utility function $U_{\text{true}}$ of the user is as follows:
\begin{equation}
    \label{eq:full_utrue_ranking}
    \begin{gathered}
        r_5 \rightarrow r_{11} \rightarrow r_3 \rightarrow r_6 \rightarrow r_9 \rightarrow r_7 \rightarrow r_{12} \rightarrow r_{10} \\
        \rightarrow r_4 \rightarrow r_{13} \rightarrow r_8 \rightarrow r_{14} \rightarrow r_{0} \rightarrow r_{1} \rightarrow r_{2}\,.
    \end{gathered}
\end{equation}

It can be observed that the inferred PRUS perfectly preserves the user's reference ranking of the 10 selected rules. However, when PRUS ranks the complete set of 16 covering rules, its ranking differs a bit from that produced by $U_{\text{true}}$. The ranking fidelity, measured by Kendall's $\tau$ between the rankings of all 16 rules produced by PRUS and $U_{\text{true}}$, equals $0.4095$. 

\begin{table}[tbp]
    \centering
    \small
    \begin{tabular}{cccccccccccccccc}
        \toprule
        & & \multicolumn{9}{c}{\textbf{Elementary conditions}} & \multicolumn{3}{c}{\textbf{Rule eval. criteria}} \\
        \cmidrule(lr){3-11} \cmidrule(lr){12-14}
        
        \textbf{Rule} & & \rot{Thick} & \rot{Size} & \rot{Shape} & \rot{Adh} & \rot{Epith} & \rot{Bare} & \rot{Chrom} & \rot{Nuc} & \rot{Mit} & Compl. & Supp. & Conf. & \textbf{PRUS} \\

        \midrule
        $r_5$  &         &   & $\bullet$ &   &   &   & $\bullet$ &   & $\bullet$ &   & 0.333 & 0.515 & 0.798 & 0.563 \\
        $r_3$  & $\star$ &   & $\bullet$ & $\bullet$ &   &   & $\bullet$ &   &   &   & 0.333 & 0.549 & 0.852 & 0.524 \\
        $r_{11}$ &         & $\bullet$ & $\bullet$ &   &   &   &   &   & $\bullet$ &   & 0.333 & 0.194 & 0.301 & 0.512 \\
        $r_6$  & $\star$ &   &   & $\bullet$ &   &   & $\bullet$ &   & $\bullet$ &   & 0.333 & 0.527 & 0.818 & 0.483 \\
        $r_9$  & $\star$ & $\bullet$ & $\bullet$ &   & $\bullet$ &   &   &   &   &   & 0.333 & 0.187 & 0.290 & 0.442 \\
        $r_7$  & $\star$ &   &   & $\bullet$ &   &   & $\bullet$ &   &   & $\bullet$ & 0.333 & 0.582 & 0.903 & 0.401 \\
        $r_{12}$ & $\star$ & $\bullet$ &   & $\bullet$ & $\bullet$ &   &   &   &   &   & 0.333 & 0.194 & 0.301 & 0.360 \\
        $r_4$  &         &   & $\bullet$ &   &   & $\bullet$ & $\bullet$ &   &   &   & 0.333 & 0.469 & 0.727 & 0.358 \\
        $r_{10}$ & $\star$ & $\bullet$ & $\bullet$ &   &   &   & $\bullet$ &   &   &   & 0.333 & 0.192 & 0.298 & 0.319 \\
        $r_8$  & $\star$ &   &   & $\bullet$ &   & $\bullet$ & $\bullet$ &   &   &   & 0.333 & 0.484 & 0.750 & 0.278 \\
        $r_{13}$ &         & $\bullet$ &   & $\bullet$ &   &   & $\bullet$ &   &   &   & 0.333 & 0.196 & 0.304 & 0.236 \\
        $r_{14}$ & $\star$ & $\bullet$ &   &   & $\bullet$ &   & $\bullet$ &   &   &   & 0.333 & 0.179 & 0.278 & 0.213 \\
        $r_0$  & $\star$ & $\bullet$ &   &   &   & $\bullet$ &   &   &   &   & 0.222 & 0.187 & 0.290 & 0.090 \\
        $r_1$  &         & $\bullet$ &   &   &   &   &   & $\bullet$ &   &   & 0.222 & 0.057 & 0.088 & 0.054 \\
        $r_2$  & $\star$ &   &   &   &   &   & $\bullet$ & $\bullet$ &   &   & 0.222 & 0.178 & 0.276 & 0.050 \\
        \bottomrule       
    \end{tabular}
    \caption{\textbf{Personalized Rule Ranking.} The complete set of candidate rules covering the analyzed instance, ordered descendingly by the derived Preference-based Rule Utility Score (PRUS). Elementary conditions in rule antecedents are indicated by the $\bullet$ symbol. The $\star$ symbol indicates rules evaluated by the user in the reference ranking. \textbf{Intervals of feature values in elementary conditions:} \textit{Thick} (Clump Thickness $\in$ 3--5), \textit{Size} (Uniformity of Cell Size $\in$ 1--3), \textit{Shape} (Uniformity of Cell Shape $\in$ 1--3.33), \textit{Adh} (Marginal Adhesion $\in$ 1--3), \textit{Epith} (Single Epithelial Cell Size $\in$ 2--3), \textit{Bare} (Bare Nuclei $\in$ 1--3), \textit{Chrom} (Bland Chromatin $\in$ 3--10), \textit{Nuc} (Normal Nucleoli $\in$ 1--2), \textit{Mit} (Mitoses $\in$ 1--10).}
    \label{tab:toy-example}
\end{table}

What is worth noting is that two new rules, which were not considered by the user in the reference ranking, appear at the top of the PRUS ranking.
These are rule $r_5$, which achieves the highest overall utility score ($0.563$),
and rule $r_{11}$, which is ranked third, immediately after the first rule in the reference ranking. This finding constitutes a clear benefit of our method. It is formally captured by the rule discovery metric, which yields a value of 2, meaning that two entirely new rules have appeared in the top-5 positions of the algorithmically generated ranking. Remarkably, the same two rules are at the top of the ground-truth ranking provided by $U_{\text{true}}$. 

To ensure these algorithmically discovered rules are of genuine interest to the user, we calculate the Jaccard index for the top-5 positions of both rankings; this yields a perfect value of 1.0, demonstrating absolute agreement between the inferred and actual top preferences. This result highlights the ability of the proposed method to mathematically generalize the user’s elicited preferences, leading to the discovery and recommendation of novel, highly relevant explanations.

\subsection{Analysis of experimental results and discussion}

This section presents an analysis of the main experimental results for all three considered datasets. The analysis is structured around the defined research questions.

\begin{figure}[tbp]
    \centering
    \includegraphics[width=\linewidth]{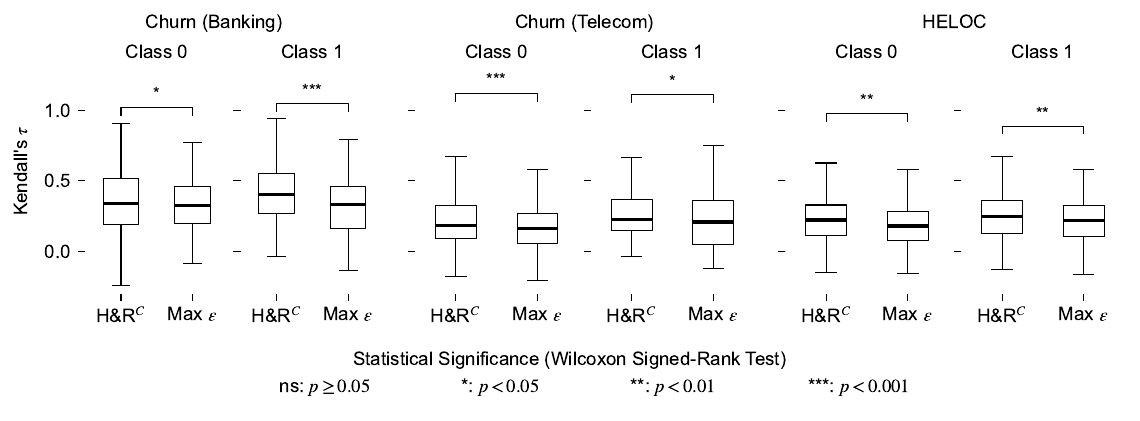}
    \caption{\textbf{Ranking Fidelity:} Distribution of Kendall's $\tau$ correlation between the algorithmically generated full rankings and the true user preferences. Higher values indicate that the algorithm better preserves the user's overall preference order.}
    \label{fig:kendall-ranking}
\end{figure}

\begin{figure}[tbp]
    \centering
    \includegraphics[width=\linewidth]{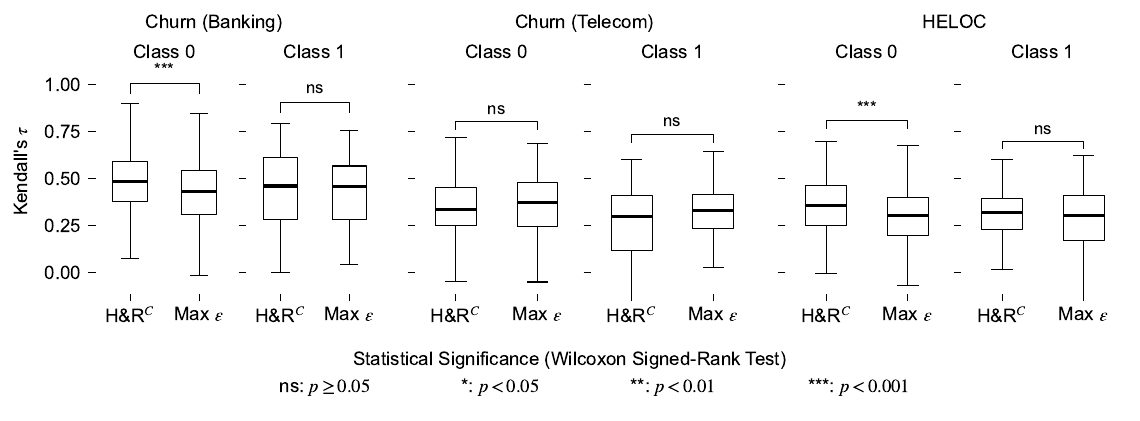}
    \caption{\textbf{Parameter Recovery:} Distribution of Kendall's $\tau$ correlation between the recovered PRUS weights and the user's ground-truth $U_{\text{true}}$ weights. Higher values demonstrate a more accurate approximation of the underlying utility function.}
    \label{fig:kendall-weights}
\end{figure}

\subsubsection{RQ1: Alignment of Top-Ranked Rules}
\begin{figure}[tbp]
    \centering
    \includegraphics[width=0.5\linewidth]{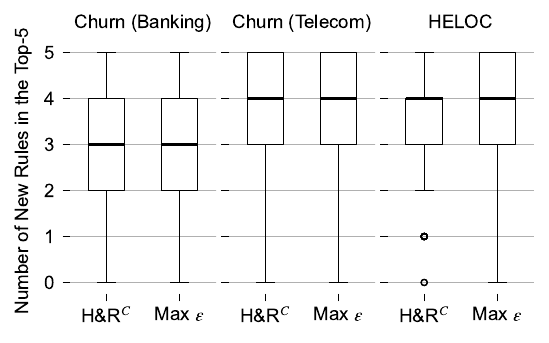}
    \caption{\textbf{Rule Discovery within Top Positions:} The number of newly discovered rules occupying the top-5 positions of the final ranking. Values range from 0 to 5, showing how many top rules in the algorithmic solution are new to the user.}
    \label{fig:number-of-discovered-rules-top-5}
\end{figure}

\begin{figure}[tbp]
    \centering
    \includegraphics[width=\linewidth]{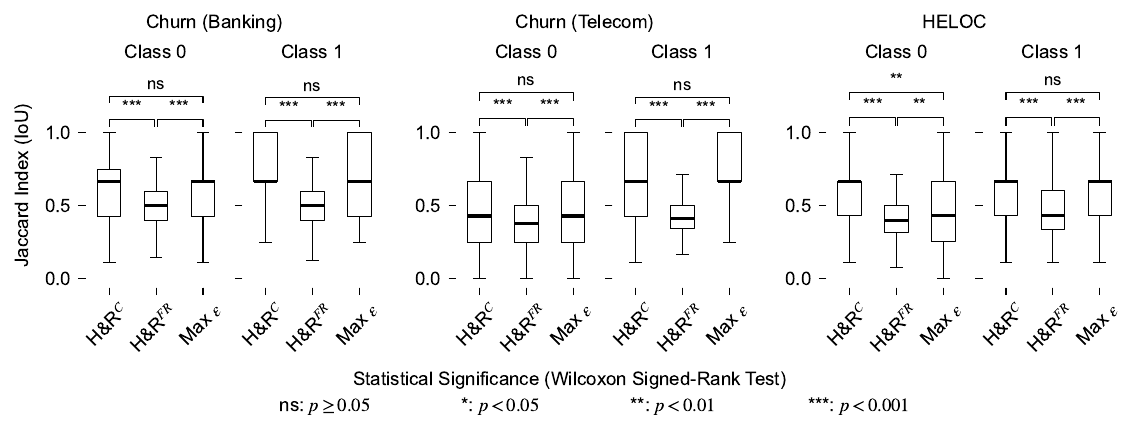}
    \caption{\textbf{Top-5 Rule Set Similarity:} Jaccard index measuring the overlap between the top-5 rules selected by the algorithm and the user's true top-5 rules. A score closer to 1.0 indicates near-perfect alignment with the most preferred rules.}
    \label{fig:jaccard-top-5}
\end{figure}

\begin{figure}[tbp]
    \centering
    \includegraphics[width=\linewidth]{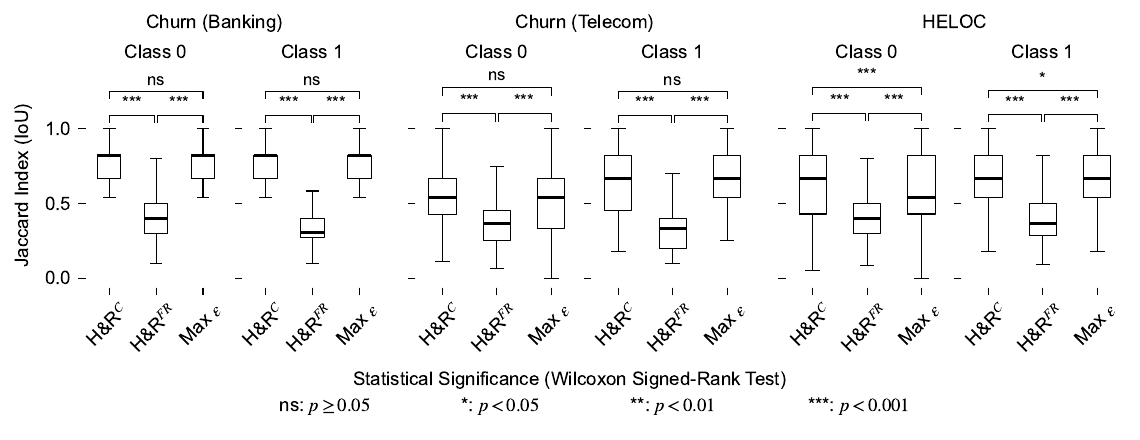}
    \caption{\textbf{Top-10 Rule Set Similarity:} Jaccard index measuring the overlap between the top-10 rules selected by the algorithm and the user's true top-10 rules. A score closer to 1.0 indicates better identification of the user's preferred explanatory patterns.}
    \label{fig:jaccard-top-10}
\end{figure}

RQ1 is evaluated using the Jaccard index for the top-5 and top-10 rules in both rankings. For the top-5 rules (Figure~\ref{fig:jaccard-top-5}), both ordinal regression algorithms, Max~$\epsilon$ and H\&R$^C$, achieve high median overlap with the user's true top rules. Conversely, the alternative H\&R$^{FR}$ (first rules) strategy yields notably lower Jaccard scores, proving less effective. This strong alignment for the proposed approaches remains consistent for the top-10 rules (Figure~\ref{fig:jaccard-top-10}).
Given its markedly lower efficacy across these foundational set similarity metrics, H\&R$^{FR}$ is demonstrably less competitive for this specific task. Consequently, to maintain focus on the most viable techniques, the remainder of our analysis concentrates exclusively on the two most prominent strategies: H\&R$^C$ and Max~$\epsilon$.
Importantly, this high overlap in the top-$k$ sets is achieved while actively introducing novel rules. As demonstrated in Figure~\ref{fig:number-of-discovered-rules-top-5}, a substantial portion of the rules occupying the final top-5 positions were absent from the initial reference ranking. This proves that the high Jaccard scores reflect a genuine generalization of the user's preferences rather than a simple memorization of the evaluated reference set.

\begin{figure}[tbp]
    \centering
    \includegraphics[width=0.5\linewidth]{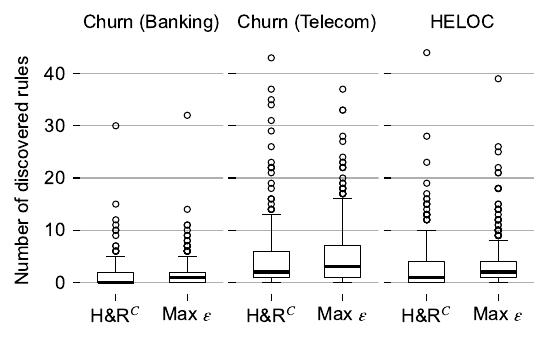}
    \caption{\textbf{Rule Discovery:} The total number of newly discovered rules in the final ranking that appear strictly before the user's highest-ranked reference rule. Higher counts indicate a stronger capacity to reveal previously unseen rules.}
    \label{fig:number-of-discovered-rules}
\end{figure}

\begin{table}[tbp]
    \centering
    \begin{tabular}{l c c}
        \toprule
        Dataset & H\&R$^{C}$ & Max $\epsilon$ \\
        \midrule
        Churn (Banking) & 1.28 & 1.66 \\
        Churn (Telecom)  & 4.60 & 5.37 \\
        HELOC           & 3.01 & 3.54 \\
        \bottomrule
    \end{tabular}
    \caption{\textbf{Mean number of rules that PRUS ranks above the reference ranking:} The average number of newly discovered rules within the final rankings appearing before the user's highest-ranked reference rule.}
    \label{tab:mean-discovered}
\end{table} 

\subsubsection{RQ2: PRUS Reproduction}
We address RQ2 by analyzing ranking fidelity and parameter recovery. Figure~\ref{fig:kendall-ranking} shows that both Max~$\epsilon$ and H\&R$^C$ achieve median Kendall's $\tau$ values indicating a positive correlation with the true ranking. Parameter recovery (Figure~\ref{fig:kendall-weights}) follows a similar trend. The correlation between recovered and ground-truth weights confirms that both algorithms successfully approximate the underlying PRUS utility functions using a limited reference ranking.

\subsubsection{RQ3: Rule Discovery Capacity}
We explicitly address RQ3 by evaluating the total number of previously unseen rules elevated to the top of the inferred rankings. As detailed in Table~\ref{tab:mean-discovered}, both optimization strategies successfully identify multiple new rules that PRUS ranks strictly above the user's highest-ranked reference rule. The full distribution of these discoveries across the evaluated datasets is visually confirmed by the boxplots in Figure~\ref{fig:number-of-discovered-rules}. These results formally substantiate the algorithm's capacity to broaden the user's explanatory perspective by recommending highly relevant, mathematically compatible rules that were not part of the initial elicitation process.

\subsubsection{RQ4: Comparison of Optimization Strategies}
Finally, we address RQ4 by comparing the hit-and-run centroid (H\&R$^{C}$) and epsilon maximization (Max $\epsilon$) strategies. H\&R$^{C}$ generally achieves slightly higher median correlations for both ranking fidelity (Figure~\ref{fig:kendall-ranking}) and parameter recovery (Figure~\ref{fig:kendall-weights}). It also performs comparably or better in top-$k$ rule set similarity (Figures~\ref{fig:jaccard-top-5} and~\ref{fig:jaccard-top-10}). Statistical tests confirm that H\&R$^{C}$ sometimes outperforms Max $\epsilon$ ($p < 0.05$), especially for class 0 instances. This suggests that expected weights from hit-and-run sampling provide a more robust PRUS approximation than a single discrimination-maximizing vector. 

Conversely, regarding rule discovery capacity, the Max $\epsilon$ strategy demonstrates a more aggressive exploratory behavior. As indicated by Table~\ref{tab:mean-discovered} and supported by Figure~\ref{fig:number-of-discovered-rules}, Max $\epsilon$ consistently elevates a higher average number of novel rules to the top positions than H\&R$^{C}$. Therefore, while H\&R$^{C}$ yields a more robust and faithful overall approximation of the underlying preferences, Max $\epsilon$ may be advantageous when the user's primary goal is to discover the maximum number of alternative explanatory rules.

In \ref{app:ref_ranking}, we investigate how the size of the initial reference ranking affects the algorithmic PRUS reproduction. The results demonstrate that increasing the ranking length---that is, requiring the user to evaluate a larger subset of rules---yields continuous improvements up to a saturation point of roughly 12 to 14 rules, depending on the metric used.

\section{Conclusion and Future Research Directions}

This paper introduces a novel methodology for personalizing local, rule-based explanations of black-box machine learning models. By integrating formal preference modeling into the explainable AI pipeline, our approach tailors exact logical rules to the specific preferences of individual users without sacrificing the predictive accuracy of the underlying model. 

Computational experiments on three real-world datasets confirmed the usefulness of this framework. The results demonstrate that applying robust ordinal regression to a small reference ranking accurately recovers the user's implicit utility function. Specifically, the hit-and-run centroid (H\&R$^C$) strategy proved highly robust, consistently identifying the user's most preferred explanations and successfully discovering novel, highly relevant rules that were absent from the initial evaluation set.

Beyond the specific method proposed in this paper, the Preference-Based Explainable Artificial Intelligence (PREF-XAI) perspective points toward a broader shift in how explanations are conceived in explainable artificial intelligence. Rather than being treated as fixed outputs, explanations can be regarded as alternatives to be evaluated and selected according to user preferences, thus opening new avenues for adaptive and decision-oriented explanation systems. In this context, our future research will focus on broadening the applicability of the proposed methodology. The framework could be extended to personalize other explainability methods, such as counterfactuals and prototypes, and adapted to data modalities beyond tabular datasets. Moreover, future work should include empirical studies involving human decision-makers to assess how personalized explanations affect trust and practical decision-making in real-world settings.

\section*{Data and Code Availability}
The source code and instructions required to reproduce the computational experiments presented in this paper are publicly available in the GitHub repository:~\href{https://github.com/jkarolczak/personalized-rule-explanations}{https://github.com/jkarolczak/personalized-rule-explanations}. All datasets utilized in this study are publicly accessible, with their respective sources referenced in the main text.

\section*{Acknowledgments} 
\noindent The research of Salvatore Greco and Roman Słowiński was funded by the National Science Centre, Poland, under the MAESTRO programme (grant number 2023/50/A/HS4/00499).
The research of Jacek Karolczak and Jerzy Stefanowski was funded by the National Science Centre, Poland, under the OPUS programme (grant number 2023/51/B/ST6/00545).

\bibliographystyle{elsarticle-num} 
\bibliography{bibliography,Full_bibliography}

\newpage

\appendix
\section{Impact of the Reference Ranking Length}
\label{app:ref_ranking}

This appendix investigates how the size of the initial reference ranking affects the performance of the proposed methodology. The reference ranking length defines the number of rules evaluated by the user, which determines the number of constraints applied to the ordinal regression problem.

\begin{figure}[h]
    \centering
    \includegraphics[width=\linewidth]{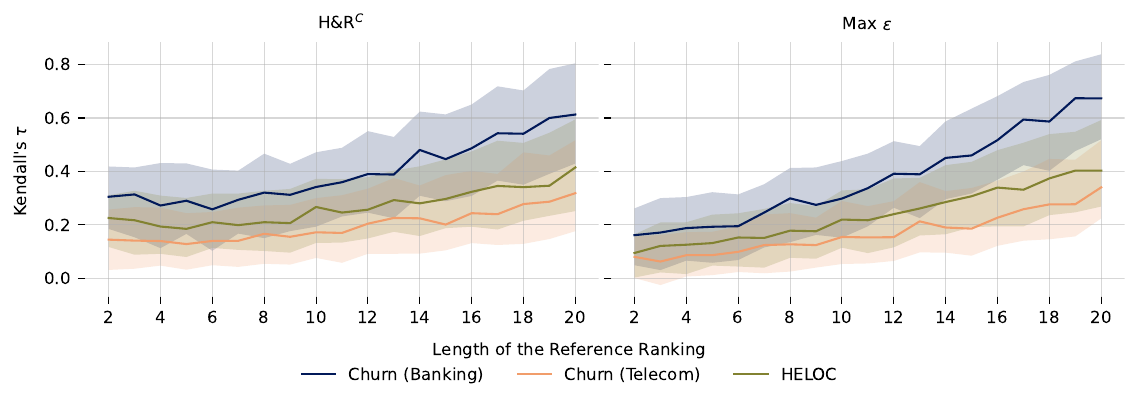}
    \caption{\textbf{Ranking Fidelity vs. Reference Ranking Length:} Distribution of Kendall's $\tau$ correlation between the algorithmically generated rankings and the simulated user's true preferences across varying lengths of the initial reference ranking.}
    \label{fig:kendall-ranking-rrl}
\end{figure}

\begin{figure}[h]
    \centering
    \includegraphics[width=\linewidth]{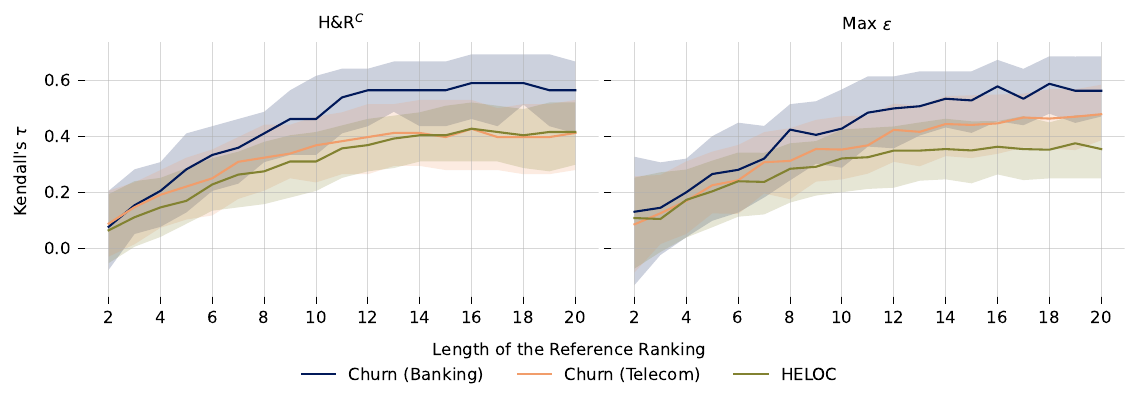}
    \caption{\textbf{Parameter Recovery vs. Reference Ranking Length:} Distribution of Kendall's $\tau$ correlation between the recovered PRUS weights and the ground-truth weights of the user, evaluated at different reference ranking lengths.}
    \label{fig:kendall-weights-rrl}
\end{figure}

Figures~\ref{fig:kendall-ranking-rrl} and \ref{fig:kendall-weights-rrl} illustrate the impact of the reference ranking length on ranking fidelity and parameter recovery, respectively. As the number of ranked rules increases, more preference information is supplied to the model. This additional context systematically narrows the space of compatible weight vectors, leading to a more accurate approximation of the underlying Preference-based Rule Utility Score (PRUS). Consequently, a steady improvement in Kendall's $\tau$ correlation is observed across the evaluated datasets.

\begin{figure}[h]
    \centering
    \includegraphics[width=\linewidth]{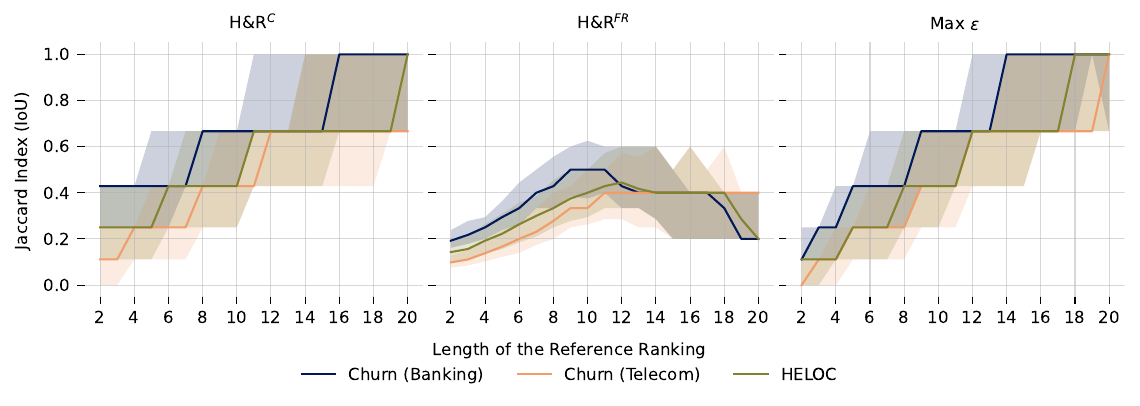}
    \caption{\textbf{Top-5 Set Similarity vs. Reference Ranking Length:} Jaccard index measuring the overlap between the algorithm's top-5 selected rules and the user's true top-5 rules. Performance is compared between the strategies as the reference length increases.}
    \label{fig:jaccard-top-5-rrl}
\end{figure}

Similarly, Figure~\ref{fig:jaccard-top-5-rrl} demonstrates that the Top-5 set similarity improves with longer reference rankings. Both the hit-and-run centroid (H\&R$^{C}$) and the first rules (H\&R$^{FR}$) strategies exhibit higher Jaccard indices as more rules are evaluated by the user.

While increasing the reference ranking length enhances the mathematical fidelity of the recovered preferences, it simultaneously imposes a higher cognitive burden on the user. The empirical results suggest that a moderate ranking length, typically between 7 and 12 rules, offers an effective compromise. This range provides sufficient constraints to accurately tailor the explanations while keeping the user's evaluation effort manageable.

\end{document}